\documentclass{sigchi}

\CopyrightYear{2020}
\setcopyright{acmlicensed}
\doi{}
\isbn{}
\conferenceinfo{UIST'20,}{Oct 20--23, 2020}
\acmPrice{}

 \toappear{
 Note: This is the author's preprint copy. This is not the official ACM published version, which will appear at UIST 2020.
 }

\usepackage{balance}       
\usepackage{graphics}      
\usepackage[T1]{fontenc}   
\usepackage{txfonts}
\usepackage{mathptmx}
\usepackage[pdflang={en-US},pdftex]{hyperref}
\usepackage{color}
\usepackage{booktabs}
\usepackage{textcomp}
\usepackage{microtype}        
\usepackage{ccicons}          

\usepackage{todonotes}

\def\plaintitle{Predicting Visual Importance Across Graphic Design Types}
\def\plainauthor{First Author, Second Author, Third Author,
  Fourth Author, Fifth Author, Sixth Author}

\def\plainkeywords{Graphic Designs; Importance; Saliency; Human Attention; Automated Design; User Interface for Design; Deep Learning}

\makeatletter
\def\url@leostyle{%
  \@ifundefined{selectfont}{
    \def\UrlFont{\sf}
  }{
    \def\UrlFont{\small\bf\ttfamily}
  }}
\makeatother
\def\pprw{8.5in}
\def\pprh{11in}

\definecolor{linkColor}{RGB}{6,125,233}
\hypersetup{%
  pdftitle={\plaintitle},
  pdfauthor={\plainauthor},
  pdfkeywords={\plainkeywords},
  pdfdisplaydoctitle=true, 
  bookmarksnumbered,
  pdfstartview={FitH},
  colorlinks,
  citecolor=black,
  filecolor=black,
  linkcolor=black,
  urlcolor=linkColor,
  breaklinks=true,
  hypertexnames=false
}

\newcommand{\cam}[1]{\textcolor{black}{#1}}
\newcommand{\zoya}[1]{\textcolor{black}{#1}}

\newcommand{\peter}[1]{\textcolor{black}{#1}}

\begin{document}

\title{\plaintitle}

\numberofauthors{1}
\author{%
 \alignauthor{
    Camilo Fosco\textsuperscript{1},
    Vincent Casser\textsuperscript{1},
    Amish Kumar Bedi\textsuperscript{2},\\
    Peter O'Donovan\textsuperscript{2},
    Aaron Hertzmann\textsuperscript{3},
    Zoya Bylinskii\textsuperscript{3}, \\
   \affaddr{\textsuperscript{1}MIT}
   \affaddr{\textsuperscript{2}Adobe Inc.}
   \affaddr{\textsuperscript{3}Adobe Research}\\
   \email{\{camilolu, vcasser\}@mit.edu}
   \email{\{ambedi, podonova, hertzman, bylinski\}@adobe.com}
   }
 }

\maketitle

\begin{abstract}
This paper introduces a Unified Model of Saliency and Importance (UMSI), which learns to predict visual importance in input graphic designs, and saliency in natural images, along with a new dataset and applications.  Previous methods for predicting saliency or visual importance are trained individually on \zoya{specialized} datasets, making them limited in application and leading to poor generalization on novel \zoya{image classes}, while requiring a user to know which model to apply to which input. UMSI is \zoya{a deep learning-based model simultaneously} trained on images from different design classes, including posters, infographics, mobile UIs, as well as natural images, and includes an automatic classification module to classify the input. This allows the model to work more effectively without requiring a user to \zoya{label} the input. We also introduce Imp1k, a new dataset of designs annotated with importance information. We demonstrate two new design interfaces that use importance prediction, including a tool for adjusting the relative importance of design elements, and a tool for reflowing designs to new aspect ratios while preserving visual importance.\footnote{The model, code, and importance dataset are available at:\\ \url{http://predimportance.mit.edu}.}
\end{abstract}

 {}{}


\section{Introduction}

Where a viewer looks on a poster or advertisement can determine whether the design is effective in communicating its message to the viewer, or if the viewer misses important concepts or details. Indeed, numerous companies offer eye tracking analyses of graphic designs, providing insight into the effectiveness of the design, but these typically depend on a separate eye tracking study for each design.  \peter{Establishing visual hierarchy and importance is a common task for designers, ideally guiding the reader through the design from elements of higher to lesser importance.} Methods that predict visual importance accurately in real-time could \zoya{thus} be used in a number of UI applications, including providing real-time feedback to designers, as well as automatic reflow. 

\zoya{Prior approaches have tackled various aspects of attention modeling on graphic designs, including saliency on visualizations~\cite{matzen2017data} and mobile UIs~\cite{gupta2018saliency}, visual flow on comics~\cite{cao2014look} and webpages~\cite{shen2014webpage}, and importance prediction on visualizations~\cite{bylinskii2017learning} and posters~\cite{bylinskii2017learning,o2014learning}. Unlike saliency or visual flow which model eye fixations and trajectories, respectively, importance identifies design elements of interest/relevance to the viewer. This makes it more practical as a building block for downstream design applications. Moreover, the narrow focus of prior work on particular design types, datasets, and tasks makes the models difficult to adapt to other problems and generalize to broader sets of design classes.}  Graphic designs in domains such as advertising, art and education show significant differences in terms of content, layout and appearance, \zoya{which} means that a designer would need to find a model trained \zoya{specifically} for their class of design.  

\begin{figure}
  \centering
  \includegraphics[width=1\columnwidth]{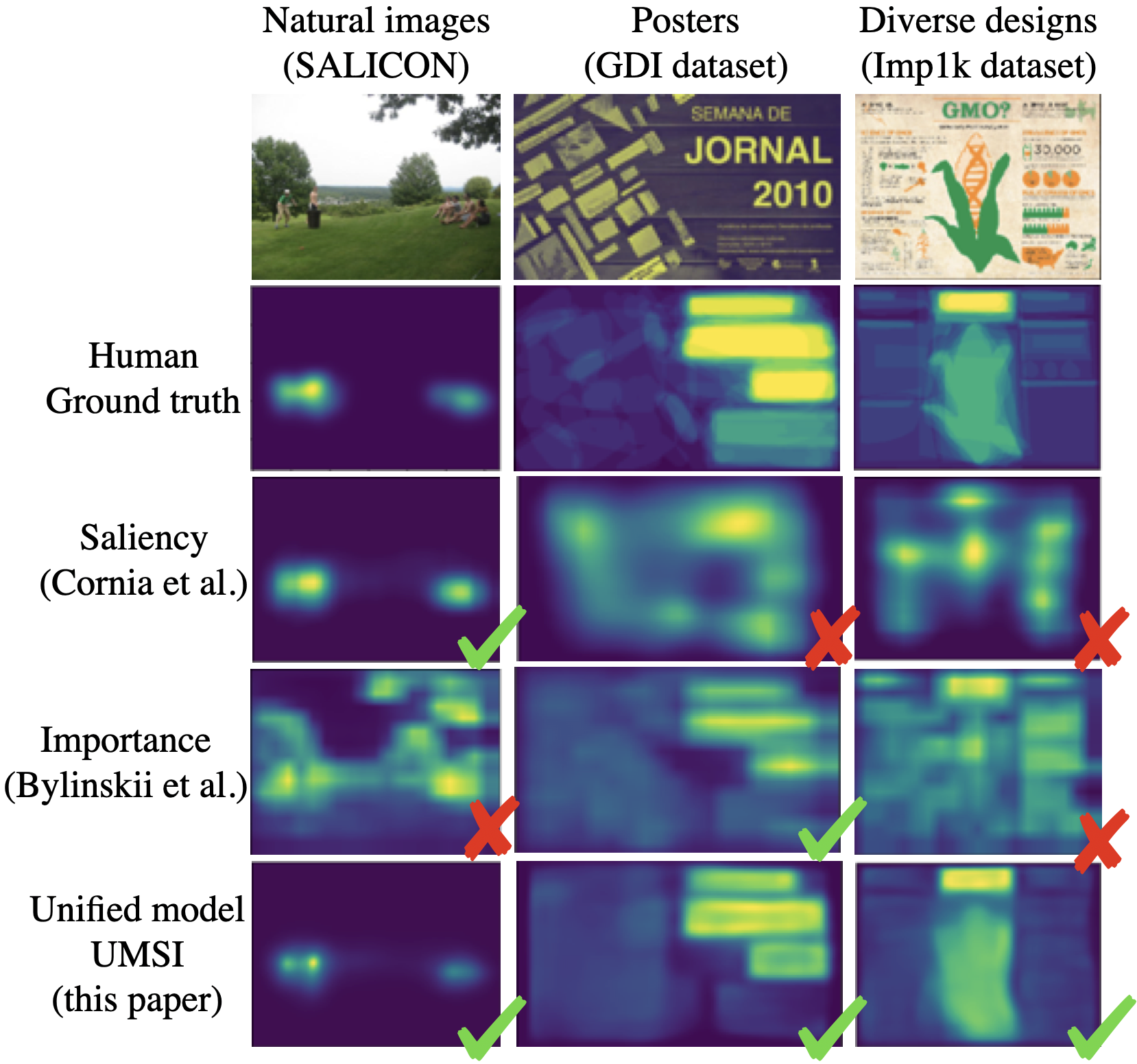}
  \caption{Our Unified Model of Saliency and Importance, UMSI, is able to predict human attention on both natural images and graphic designs. Its performance is comparable to state-of-the-art models specifically built for natural image saliency~\protect\cite{cornia2018predicting} and it outperforms the existing visual importance models~\protect\cite{bylinskii2017learning,o2014learning}.}  
  \label{sam_vs_byl_vs_UMSI}
\end{figure}

This paper proposes a unified model for predicting importance in graphic designs from multiple design classes, as well as for predicting saliency in natural images.  Our model, the Unified Model of Saliency and Importance (UMSI), is a deep neural network that automatically classifies the input design into one of five design classes before predicting the importance or saliency of the input image. Moreover, we introduce a new dataset, Imp1k, that contains 1000 annotated designs, covering webpages, movie posters, mobile UIs, infographics and advertisements.
We show that our model either outperforms or remains competitive with state-of-the-art approaches on previous datasets of graphic designs and natural images (Figure \ref{sam_vs_byl_vs_UMSI}). We also show that current state-of-the-art models struggle on Imp1k, highlighting a lack of generalization to diverse design classes.  Our model successfully generalizes to the attention patterns that are unique to these different design classes. 

This approach offers several benefits.  First, our model can take many different kinds of designs and images as input, without having to train a separate model for each.  A user does not need to \zoya{explicitly label} or classify the input design. Second, the model leverages training data more effectively by using a shared representation, learning what is both common and unique across datasets. Indeed, we show that our unified model gives superior performance on individual tasks as compared to models trained on \zoya{individual} tasks. 
Third, design represents a spectrum, and our model can generalize \zoya{accordingly}. 
Since our model trains on both designs and natural images, it works well on designs that include photographs, whether the photograph dominates the design or represents a small portion of it.
While our method is already suitable for real applications, our training procedure can be generalized to more design categories in the future, simply by adding more classes.

Furthermore, we introduce two new interactive applications enabled by this model. In our first application, a designer can specify the target importance of an element in an existing design, and the algorithm updates the layout to achieve this goal. Second, we show a reflow application that converts a vector design into a new size and aspect ratio while maintaining the relative importance of design elements. Reflow is a critical problem for modern designers, who are typically tasked to retarget designs to a wide variety of display sizes and form factors. We show that using UMSI in these applications gives substantially better results than with existing baselines.

To summarize, our contributions include: UMSI, a unified model for predicting importance in different kinds of designs, and saliency in natural images; the Imp1k dataset, containing importance annotations for 1000 designs from 5 classes; a tool for revising a design to match target importance values; and a reflow application that uses UMSI to preserve importance.

\section{Related Work}

Many previous studies analyze design perception, understanding, and memorability, e.g., \cite{borkin2016beyond,borkin2013makes,bylinskii2015eye, harrison2015infographic}, including the use of eye tracking as an analysis tool. However, obtaining reliable eye movement data is typically far too costly and time-consuming to be used in most design scenarios.

Most prior work on automatic saliency prediction has focused on eye movements in natural images. Lately, these efforts have progressed significantly with the help of deep learning \cite{bylinskii2015saliency, cornia2016deep, cornia2018predicting,jiang2015salicon, kummerer2014deep, kummerer2016deepgaze}.  
When applied to graphic designs, models \zoya{trained on natural images} perform poorly \cite{bylinskii2017learning,haass2016modeling}.
\zoya{Specialized saliency models have been designed for visualizations~\cite{matzen2017data}, mobile UIs~\cite{cao2014look} and webpages~\cite{shen2014webpage,zheng2018task}, but with limited generalization ability outside each design class.}

\zoya{In contrast to saliency,} O'Donovan et al.\ \cite{o2014learning} \zoya{introduced} the first approach using importance: they collected the GDI dataset, a crowdsourced dataset where people were asked to manually label what they thought was important in a design. With this data, they trained a model for importance prediction on graphic designs. Their method, however, requires annotation of design element position and alignment to generalize to new designs. Bylinskii et al. \cite{bylinskii2017learning} followed up with the first end-to-end deep learning approach for importance prediction. Their model predicts importance from the original image using a Fully Convolutional Network \cite{long2015fully}, with pre-trained weights from semantic segmentation, fine-tuned on the GDI dataset. 
\zoya{As in prior work \cite{bylinskii2017learning,o2014learning}, we are using a viewer-identified notion of importance rather than a designer-specified notion. We are not explicitly taking into account aesthetics or design heuristics, but are focused on modeling the behavior of the viewer of the content.}

Importance and saliency have previously been used for tasks such as retargeting \cite{bylinskii2017learning, o2014learning} and thumbnailing \cite{jiao2010visual, teevan2009visual}. For the design process, however, recent work focused on visualizing predicted importance as passive feedback for the designer \cite{bylinskii2017learning}. 
In this work, we 
introduce applications of our model within interactive interfaces where the user can more actively engage with the importance model during the design process. 
\zoya{Furthermore, our retargeting application works on vector designs, and is thus more practical for design applications, compared to image-based retargeting \cite{bylinskii2017learning}, and without requiring manual annotations as in prior work \cite{o2014learning}}.

\section{Imp1k dataset} \label{sec:impdatasets}

The previously available graphic design dataset with importance annotations is limited to posters and advertisements~\cite{o2014learning}. However, the structure of other types of designs like webpages and mobile UIs is quite different, and a model trained to predict importance on posters alone will not generalize. 
At the same time, there is great interest in predicting attention patterns on webpages and mobile UIs due to their widespread use and commercial impacts. Towards expanding the generalizability of an importance model, we collected a new graphic design dataset covering five diverse design classes and use cases: infographics representing design for knowledge transfer, webpages and mobile UIs representing design for utility, and advertisements and movie posters representing design for promotion. \zoya{While prior work addresses importance for data visualizations~\cite{bylinskii2017learning}, we omit them because their highly structured layouts (axes, data marks, etc.) make them less amenable to the design applications that are the focus of this work.} The details of the data collection and annotation are provided below, along with an analysis of the importance patterns common to, and differing between, the design classes.

\subsection{Dataset collection}

\begin{figure}
  \centering
  \includegraphics[width=0.95\columnwidth]{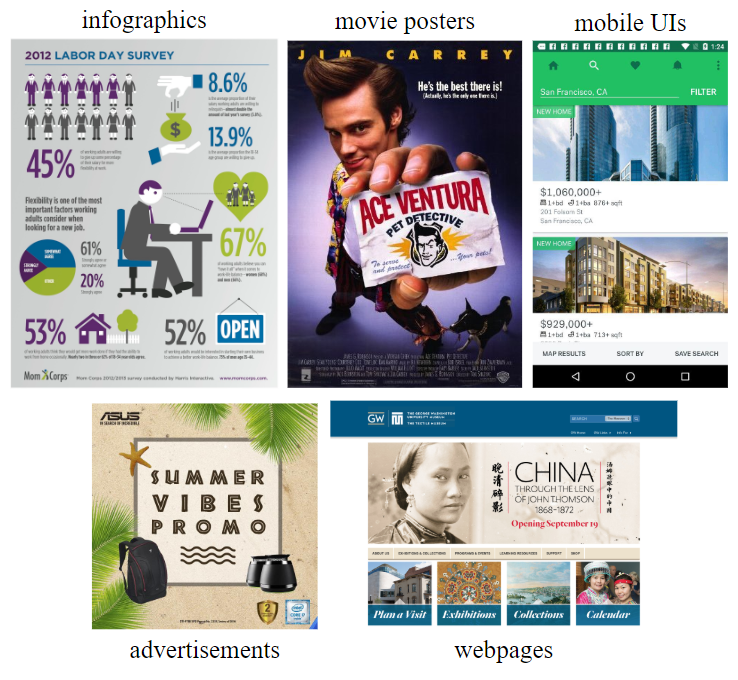}
  \caption{Examples of the 5 classes of the Imp1k dataset.} \label{fig:Imp1k}
\end{figure}

We collected designs from existing research datasets so that the stimuli and annotations may be shared broadly with the research community. Infographics were sourced from Visually29k \cite{visually1}, webpages from ClueWeb09 \cite{callan2009clueweb09}, movie posters from Chu et al.'s movie poster dataset \cite{chu2017movie}, mobile UIs from RICO \cite{deka2017rico}, and advertisements from the Pitt Image Ads \cite{hussain2017automatic}. To compose our multi-class importance dataset, Imp1k (Fig.~\ref{fig:Imp1k}), we \zoya{randomly} sampled 200 designs per design class, \zoya{after filtering out designs that had too few elements, skewed aspect ratios, or that were outliers among their design class}.

For annotating the importance in each of the 1000 designs from our Imp1k dataset, we used the ImportAnnots UI~\cite{newman2020turkeyes}. Based on the original implementation of O'Donovan et al.~\cite{o2014learning}, this UI provides participants with three options for annotation: (i)~"regular stroke" involves painting with paintbrush-like strokes with the mouse button held down, (ii) "polygon stroke" allows drawing polygons through connecting vertices, and (iii) "fill stroke" fills an area delimited by the initial mouse click position and the most recent mouse position, allowing smooth tracing over continuous contours of a shape. Using any of these tools, a participant can produce a binary mask to annotate the most important design elements. 

We deployed the ImportAnnots UI on Amazon's Mechanical Turk (MTurk) to collect importance data on all 1000 designs, splitting them into 10 designs of the same class per task (HIT), and requesting 30 participants per HIT. \cam{Participants could complete several HITs. Data from 249 individual participants was collected.} To ensure good data quality, we additionally included 3 sentinels per HIT. These are simple, artificial designs where the importance annotation is expected to cover the only visible element in the design. We designed 40 such sentinels, which we manually annotated with ground truth importance. For the task to proceed, we automatically verified that participant annotations matched the ground truth annotations (with a computed intersection over union value over 0.6) for at least 2/3 of the sentinels. Participants were paid \$1.00 for completing the task. For each of the 1000 designs collected, we computed an associated importance map by averaging the 25-30 individual binary annotations. While each participant's importance annotation is subjective, prior work has reported that the average over participants produces heatmaps that approximate attention maps~\cite{bylinskii2017learning,kim2017bubbleview,o2014learning}.

\subsection{Analysis}
\begin{figure}
\centering
  \includegraphics[width=1\linewidth]{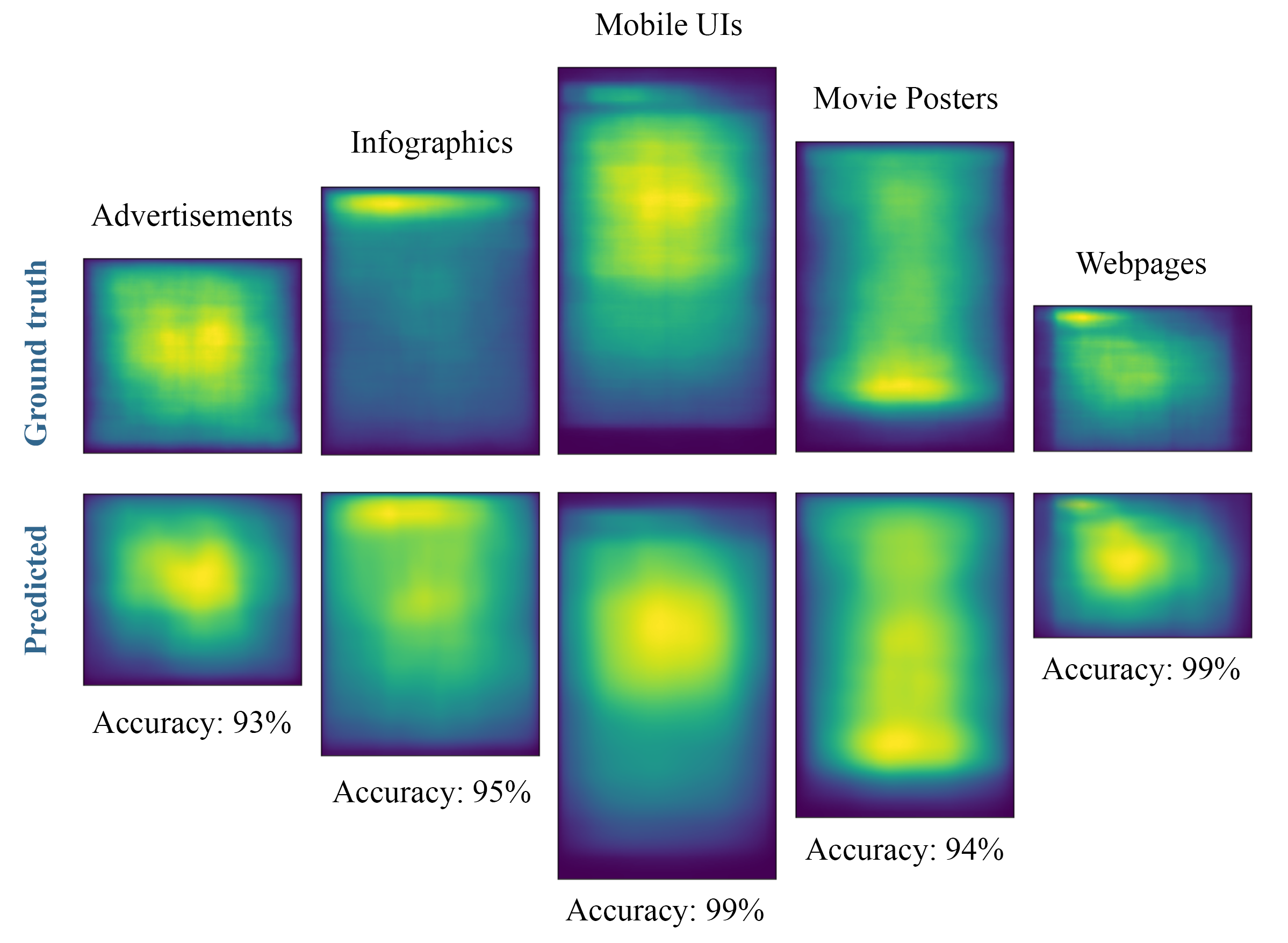}
  \caption{General importance trends on Imp1k across ground truth maps and our predicted maps. Our model captures the trends correctly, evidencing a pronounced class specialization. This is further supported by UMSI's high classification accuracy: our model learns to recognize the input's class and tailor its output to the particularities of that class.}~\label{fig:mean_trends}
\end{figure}

Each class in our dataset exhibits slightly modified patterns of importance (Fig.~\ref{fig:mean_trends}). We qualitatively observe that importance on movie posters is dominated by the title and human faces, while webpages tend to draw attention to the site name or company running the site, usually on the top left. Infographics have importance distributed across the full design, as do mobile UIs. Ads, on the other hand, concentrate most of their importance on a few elements in the center of the design. Across classes, human annotations tend to highlight a set of up to 7 different elements; more detailed annotations were rare in our data. 
We performed a basic analysis of text and face importance on different design types. \zoya{Specifically, we used open source face \cite{liu2016ssd} and text \cite{zhou2017east} detectors and calculated the mean importance map value over the face and text bounding boxes per design.} We observe that the relative importance of faces is highest on ads and movie posters, where they are nearly tied in importance. The importance of text is highest in ads, where it is perceived to be on average 32\% more important than the text in infographics (the class with the second highest average importance of text). These differences in importance patterns across classes motivate the development of a model that can generalize to these designs, and tailor predictions for each design class.

\section{UMSI Model}

Motivated by these observations, we introduce a unified importance prediction model that is trained jointly on different design classes as well as natural images. The model contains a classification branch that infers the input design category, allowing it to produce results appropriate for the input design.

\subsection{Model architecture}
\begin{figure*}
  \centering
  \includegraphics[width=1.8\columnwidth]{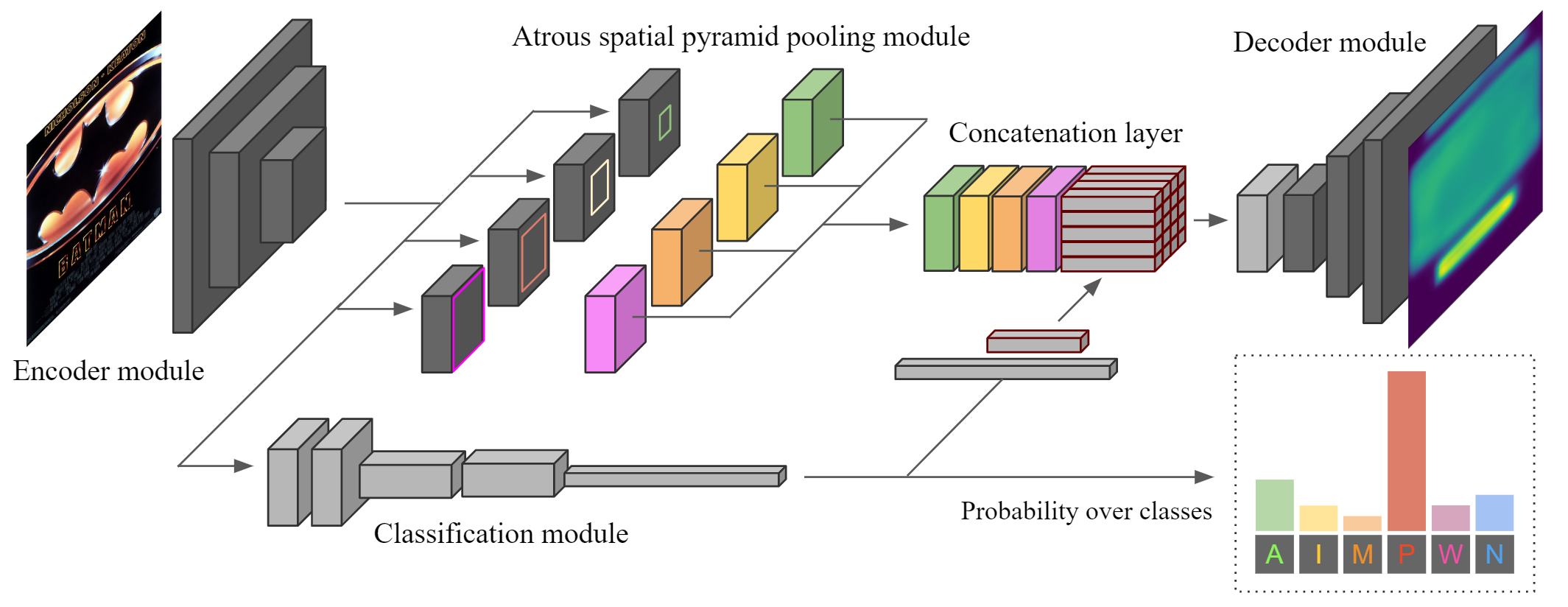}
  \caption{The architecture of our Unified Model of Saliency and Importance (UMSI). Our model leverages Xception-based encoder and decoder modules to generate accurate maps with few parameters. An atrous spatial pyramid pooling (ASPP) module captures features at different scales, and a separate classification module classifies the input into one of six categories, represented by colored letters in the figure: Ads (A), Infographics (I), Mobile-UIs (M), Movie posters (P), Webpages (W) and Natural Images (N). Central to our model is the concatenation layer, which combines the outputs from the ASPP with class-specific features obtained from the classification module and allows for tailored heatmaps for each input class.}  \label{model_arch}
\end{figure*}

Given an input image, our objective is to predict a heatmap with an importance value assigned to every pixel. This is related to other image-to-image prediction tasks like saliency and segmentation, which motivate our architectural choices. Encoder-decoder architectures are common for such tasks, whereby features are first downsampled (encoded) before being upsampled (decoded) to produce the final prediction at the original image size.
The previous state-of-the-art importance model \cite{bylinskii2017learning} was based on a semantic segmentation architecture \cite{long2015fully}. Similarly, we use a segmentation-inspired pyramid pooling module to leverage features at different scales and improve the fidelity of the heatmaps. The addition of a classification branch helps our importance model make use of class-specific information. The full architecture is visualized in Fig. \ref{model_arch} and the technical details of each component are detailed below.

\textbf{Encoder.} 
This is a low-level feature network that extracts basic image features. 
We used the Xception encoder, a high-performing state-of-the-art network common for segmentation~\cite{chen2017rethinking} and saliency~\cite{multiduration} tasks. It is composed of  depthwise separable convolutions, which 
come with great computational savings at a minimal cost on accuracy. We modified this network to output feature maps by removing the final global average pooling and fully-connected layers. To obtain higher resolution maps, 
we reduced the strides of both the 1x1 residual connection and the last two max pooling layers to one. 

\textbf{Atrous Separable Pyramid Pooling (ASPP).} To aggregate image information at multiple scales, we used an ASPP module~\cite{chen2017rethinking} after our encoding module.
This module applies several dilated convolutions with different dilation rates in parallel to a given set of feature maps. The output maps are then concatenated and passed through a projecting convolutional layer to reduce the channel dimensionality of the feature block. 
We used an ASPP module with 4 parallel convolutional layers with 256 filters, with dilation rates of 1, 6, 12 and 18. 

\textbf{Decoder.} Our decoder is a set of convolutions followed by upsampling layers and dropout. This allows the feature maps to be \zoya{scaled up} to the original image size, without a significant increase in parameters. We use three different blocks: the first two composed of two convolutional layers followed by a 2x2 upsampling layer, and the last composed of one convolutional and one upsampling layer. A 1x1 convolutional layer then transforms the feature maps into our final importance map. The convolutional layers have kernels of size 3 and the dropout applied is 0.3. We use ReLU activations throughout.

\textbf{Classification.} The output of our encoder is also passed in parallel to a classification submodule. It is composed of two 3x3 convolutional layers, followed by a convolutional layer with stride 3 to reduce dimensionality, then another 3x3 convolutional layer. The resulting feature maps are globally pooled and a dense layer with dropout transforms the maps into a 1x1x256 feature vector $C_f$. This vector is used both (i) for directly outputting the predicted image class, and (ii) for introducing class-specific information for importance prediction. In the first case, $C_f$ is fed to a dense layer with 6 outputs and softmax activation, to produce a 1x6 vector of class probabilities. In the second case, $C_f$ is resized to the required dimensions and concatenated with the output from the ASPP module along the channel dimension as part of the Concatenation layer.

\subsection{Model training}

Our challenge is to produce a model with good performance both on natural images and graphic designs, which tend to have quite different \zoya{layouts and} attributes (e.g., graphic elements, text regions, etc.). Moreover, the available ground truth attention data on natural images and graphic designs has been collected using different means, producing differences in formats \zoya{and data quantity}. The largest available dataset commonly used for training saliency models on natural images is SALICON, containing 20K images with mouse movements aggregated into saliency maps~\cite{jiang2015salicon}. The only available importance dataset for graphic designs until this paper was GDI~\cite{o2014learning}, containing 1000 designs with binary annotations aggregated into importance maps. Our Imp1k dataset also contains 1000 designs, but \zoya{split across} 5 distinct classes, while GDI is composed mainly of ads and posters from Flickr. Training a model to learn the attention patterns on both natural images and graphic designs, while using differently sized and formatted datasets, involved a specialized procedure, detailed below.  

\textbf{Training procedure.} We first pre-train our model on SALICON for 10-15 epochs. This allows our model to learn about image features broadly relevant to saliency. We then fine-tune on a graphic design importance dataset - which is either GDI or Imp1k, depending on the particular evaluation (next section). To prevent the network from ``forgetting" saliency while learning design importance, we mix in 160 new SALICON images during each fine-tuning epoch. This number maintains class balance when training on Imp1k, since the training set consists of 80\% of 200 designs per class. 

\textbf{Training details.} We use the 10K training, 5K validation, and 5K test splits from SALICON~\cite{jiang2015salicon}. For GDI, we use the same training/testing split as in \cite{bylinskii2017learning}. For Imp1k, we define a test set by randomly selecting 20\% of images from each class. 
We train with KL and CC losses, with coefficients of 10 and -3, respectively. A binary cross-entropy loss with a weight of 5 is used for the classification submodule. The learning rate is $1e^{-4}$, and is decreased by a factor of 10 every 3 epochs. To limit overfitting, we use a dropout of 0.3 on all layers. We train with a batch size of 8 using the Adam optimizer. 

\section{Model Evaluation}

We next compare with existing methods, on our dataset as well as previous importance datasets. We compare different variants, including retraining previous models on our data, and training subsets of our model (e.g., without classification), in order to evaluate each element of our method separately.


\subsection{Importance prediction}
We evaluated our model on two importance datasets, GDI and Imp1k, the results of which can be found in Tables~\ref{tab:gdi} and \ref{tab:Imp1k}, respectively. On both datasets we compare against, and outperform, the prior state-of-the-art importance model~\cite{bylinskii2017learning} across all four evaluation metrics. On the GDI dataset, UMSI notably achieves an improvement of 54\% over the prior model according to the Pearson's Correlation Coefficient (CC metric), the recommended metric for saliency evaluation~\cite{bylinskii2019different}. 

\begin{table}[h!]
\centering
\begin{tabular}{l|cccc}
\toprule
 & \multicolumn{1}{l}{\textit{$R^2$ $\uparrow$}} & \multicolumn{1}{l}{\textit{RMSE $\downarrow$}} & \multicolumn{1}{l}{\textit{CC  $\uparrow$}} & \multicolumn{1}{l|}{\textit{KL $\downarrow$}}  \\ \hline
\multicolumn{1}{l|}{O. et al., Auto \cite{o2014learning}} &  0.539 & 0.212 & - & -  \\ 
\multicolumn{1}{l|}{O. et al., Full \cite{o2014learning}} &  0.754 & 0.155 & - & -  \\ 
\multicolumn{1}{l|}{B. et al. \cite{bylinskii2017learning}} &  0.576 & 0.203 & 0.596 & 0.409  \\ 
\multicolumn{1}{l|}{B. et al. + O. \cite{bylinskii2017learning}} &  0.769 & 0.150 & - & -  \\ 
\multicolumn{1}{l|}{SAM \cite{cornia2018predicting}} &  0.693 & 0.146 & 0.884 & 0.102   \\ 
\multicolumn{1}{l|}{UMSI} & \textbf{0.781} & \textbf{0.120} & \textbf{0.915} & \textbf{0.086}                   \\ 
\end{tabular}
\caption{Evaluation of models on importance prediction on the GDI dataset. We compare our model, UMSI, to the prior state-of-the-art importance models~\protect\cite{bylinskii2017learning} and~\protect\cite{o2014learning}, and to a strong saliency model finetuned on importance~\protect\cite{cornia2018predicting}. Our model outperforms all other alternatives.}~\label{tab:gdi}
\end{table}


\begin{table}[h!]
\centering
\begin{tabular}{l|ccccc}
\toprule
 & \multicolumn{1}{l}{\textit{$R^2$ $\uparrow$}} & \multicolumn{1}{l}{\textit{RMSE $\downarrow$}} & \multicolumn{1}{l}{\textit{CC  $\uparrow$}} & 
 \multicolumn{1}{l}{\textit{KL $\downarrow$}} &
 \multicolumn{1}{l|}{\textit{Acc $\uparrow$}} \\ \hline
 \multicolumn{1}{l|}{B. et al. \cite{bylinskii2017learning}} &  0.072 & 0.181 & 0.758 & 0.301 & - \\ 
\multicolumn{1}{l|}{B. et al. x5 \cite{bylinskii2017learning}} &  0.061 & 0.205 & 0.732 & 0.388 & - \\ 
\multicolumn{1}{l|}{SAM \cite{cornia2018predicting}} &  0.108 & 0.168 & 0.866 & 0.166 & -   \\ 
\multicolumn{1}{l|}{UMSI-nc} &  0.095 & 0.152 & 0.802 & 0.177 & -  \\ 

\multicolumn{1}{l|}{UMSI-2stream} &  0.105 & 0.141 & 0.852 & 0.168 & 0.91  \\ 

\multicolumn{1}{l|}{UMSI} & \textbf{0.115} & \textbf{0.134} & \textbf{0.875} & \textbf{0.164} & \textbf{0.98}                  \\ 
\end{tabular}
\caption{Evaluation of models on importance prediction on the Imp1k dataset. We compare our model, UMSI, to the prior state-of-the-art model trained on the entire dataset (B. et al.~\protect\cite{bylinskii2017learning}) and on each class separately (B. et al. x5~\protect\cite{bylinskii2017learning}). We include SAM~\protect\cite{cornia2018predicting}, a strong saliency model also fine-tuned on the Imp1k dataset. UMSI-nc corresponds to  UMSI without a classification module, while UMSI-2stream is an alternative architecture with classification. UMSI outperforms all other alternatives.}~\label{tab:Imp1k}
\end{table}

We additionally surpass the performance of the "Full" O'Donovan model~\cite{o2014learning}, a non-automatic approach which requires human annotations of the graphical element locations at test-time. The poorer-performing, fully-automatic O'Donovan model is also included for comparison. The best-performing model out of the alternatives was a combination of the two prior models~\cite{bylinskii2019different} and \cite{o2014learning}, which was previously reported in~\cite{bylinskii2019different}. UMSI outperformed this combined model too. 

As \zoya{another} comparison point, we trained the state-of-the-art saliency model SAM~\cite{cornia2018predicting} to predict importance, by first pre-training on SALICON, and then fine-tuning on the GDI and Imp1k datasets, to report performances on each dataset. On the Imp1k dataset, we also re-trained five instances of the prior state-of-art importance model~\cite{bylinskii2017learning} on each of the 5 design classes, separately ("B. et al. x5"). From Tables~\ref{tab:gdi} and~\ref{tab:Imp1k}, we see that UMSI outperforms these alternative re-trained models, demonstrating the contributions of our architecture design, beyond the training data alone. 

\begin{figure}
\centering
  \includegraphics[width=1\linewidth]{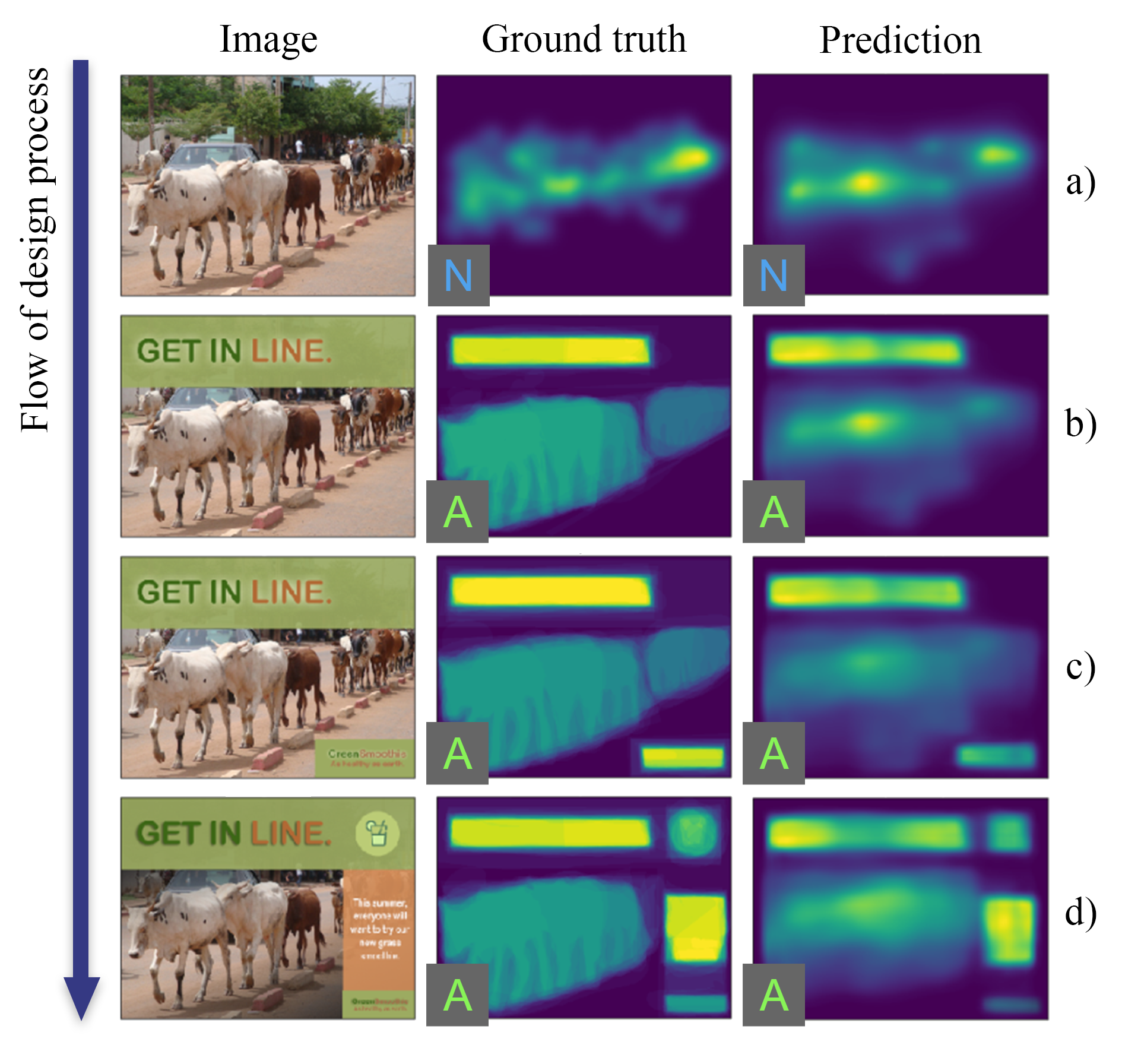}
  \caption{Our model updates importance predictions throughout the design process. Letters on the lower left represent ground truth and predicted design classes. (a) On a background natural image, UMSI predicts traditional saliency. This can inform the designer about which parts to emphasize or de-emphasize. (b) The designer adds a title element to the image. The model predicts the design is an advertisement and that the title is now most important. (c-d) As the designer adds more elements, the output map correctly captures their importance and diminishes the importance of the image background.}~\label{fig:flow_of_design}
\end{figure}

\begin{figure}
\centering
  \includegraphics[width=0.9\linewidth]{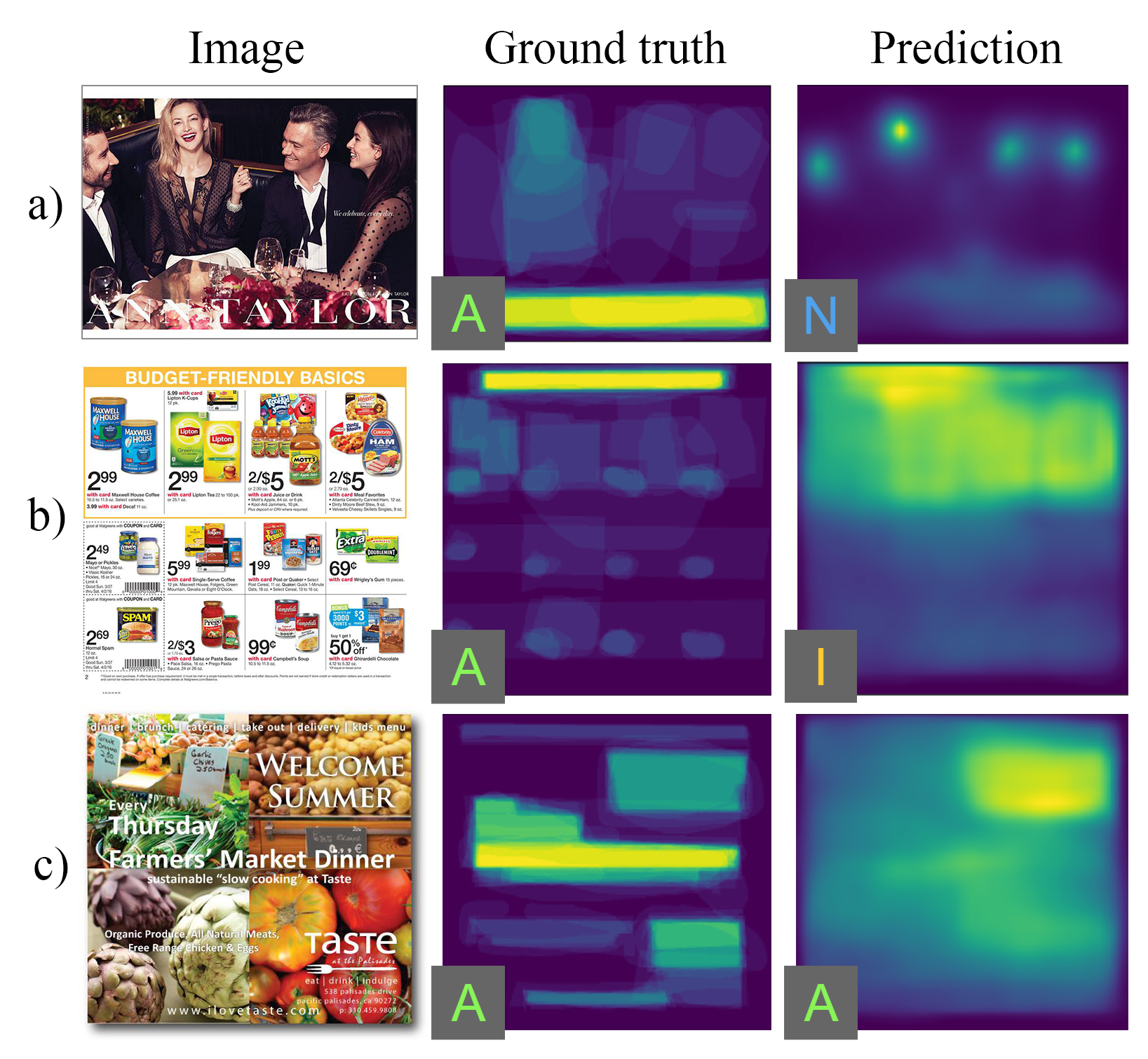}
  \caption{Failure modes of UMSI. Letters on the lower left represent ground truth and predicted design classes. (a) Our model classifies the input as a natural image, missing the subtle text overlay, and over-emphasizing the faces. (b) Our model missclassifies this ad as an infographic due to the multitude of details. Nevertheless, importance is still highest at the top of the design. (c) A complex and busy design poses a challenge for our model. Although the classification is correct, the model selects a relevant but secondary text overlay as most important.}~\label{fig:missclass}
\end{figure}

\begin{figure*}
\centering
  (a)\includegraphics[width=0.7\linewidth]{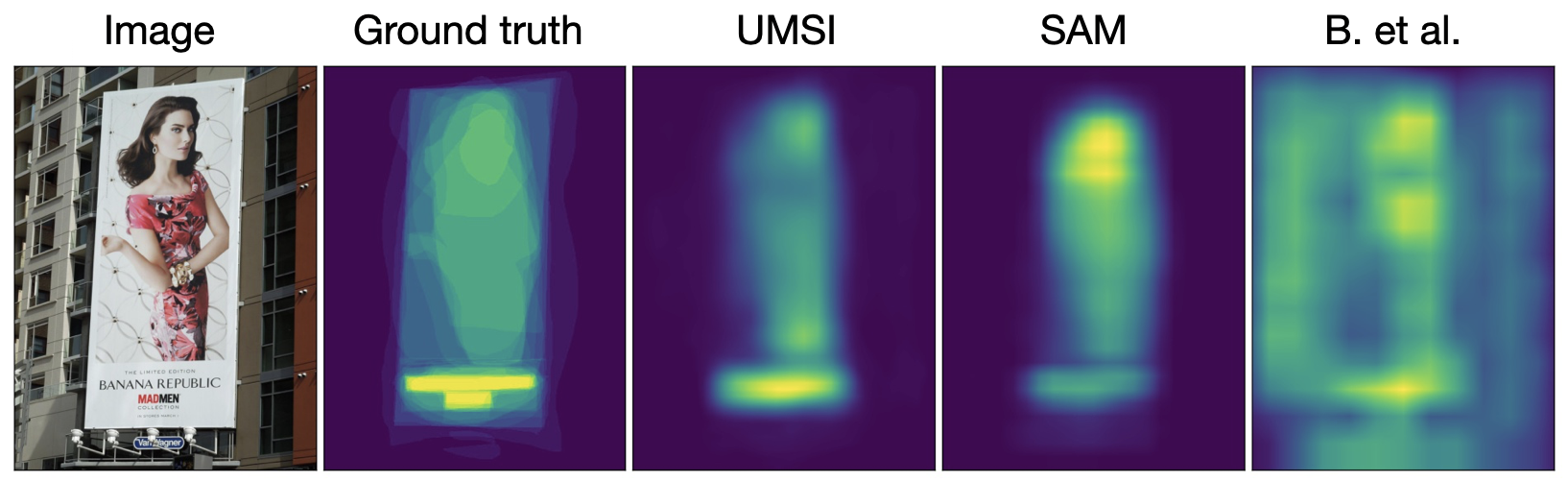}
  (b)\includegraphics[width=0.7\linewidth]{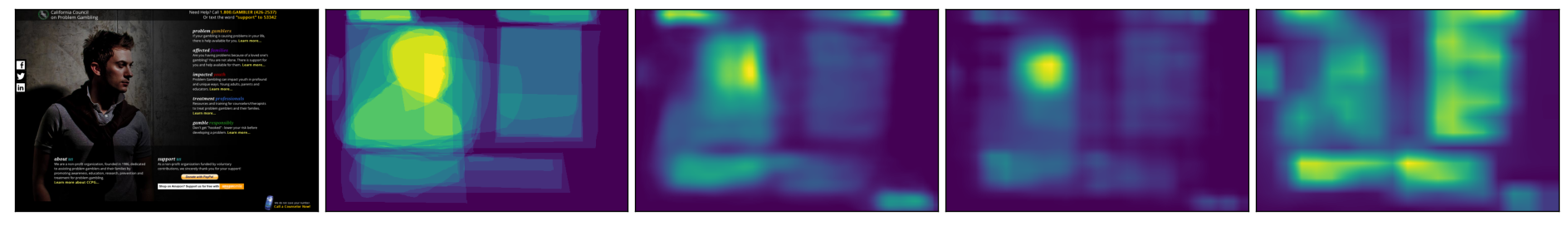}
  (c)\includegraphics[width=0.7\linewidth]{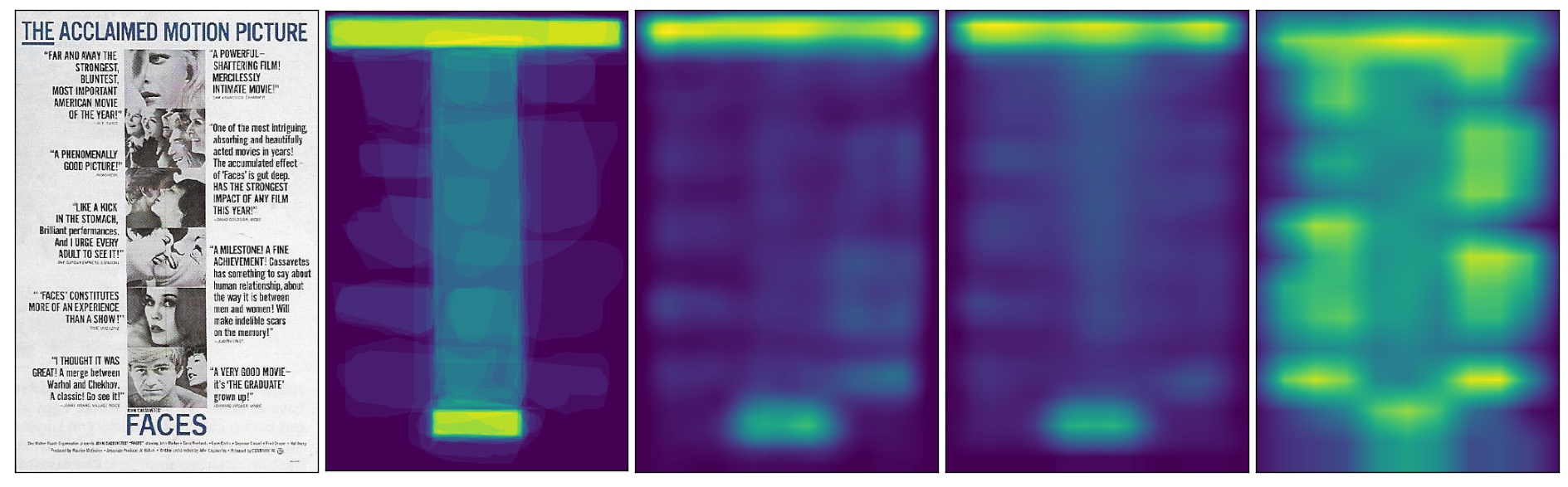}
  (d)\includegraphics[width=0.7\linewidth]{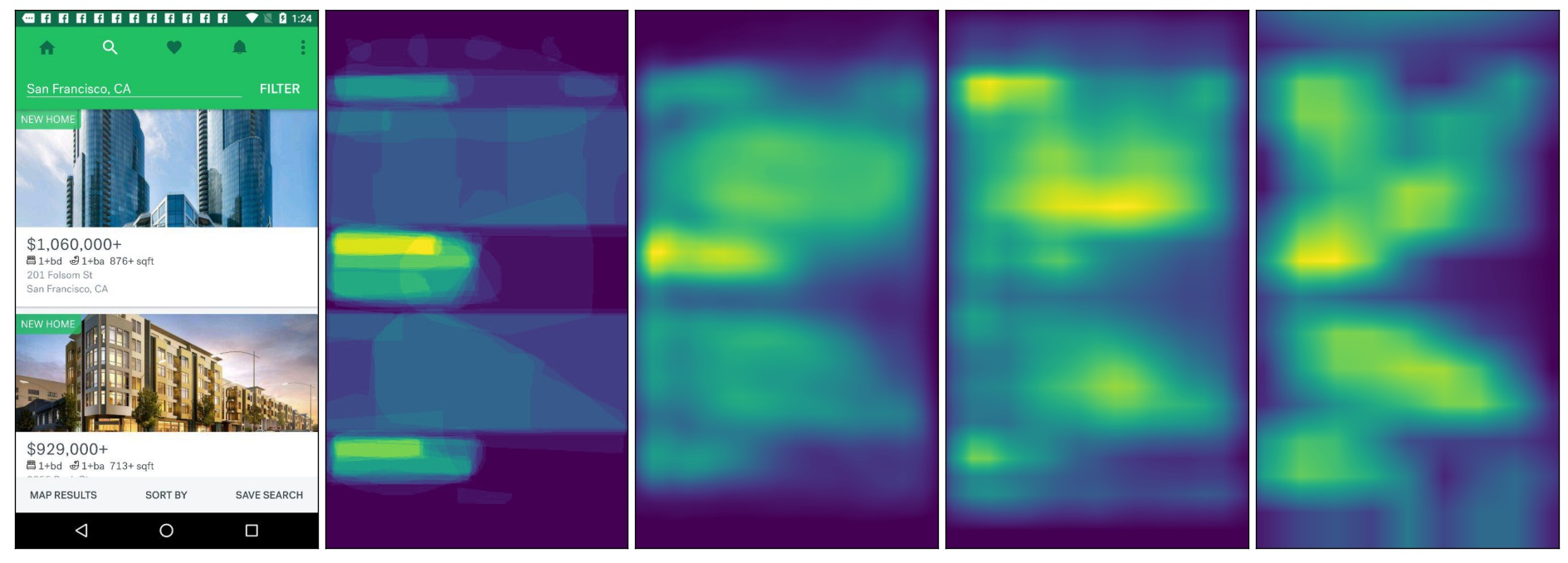}
  (e)\includegraphics[width=0.7\linewidth]{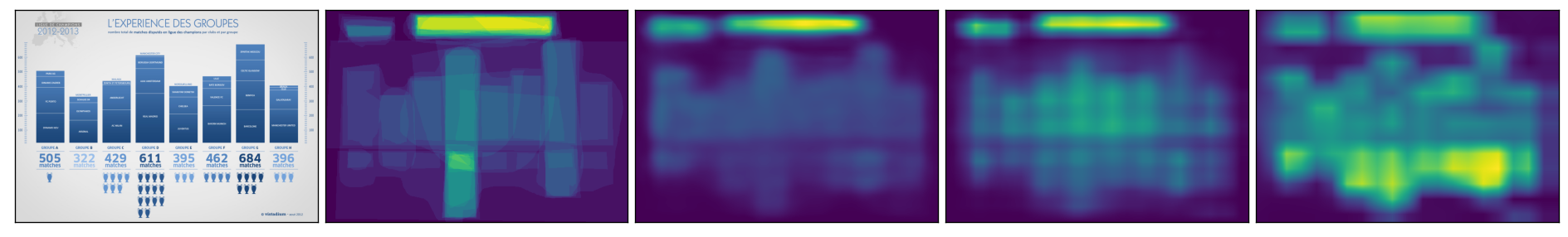}
  (f)\includegraphics[width=0.7\linewidth]{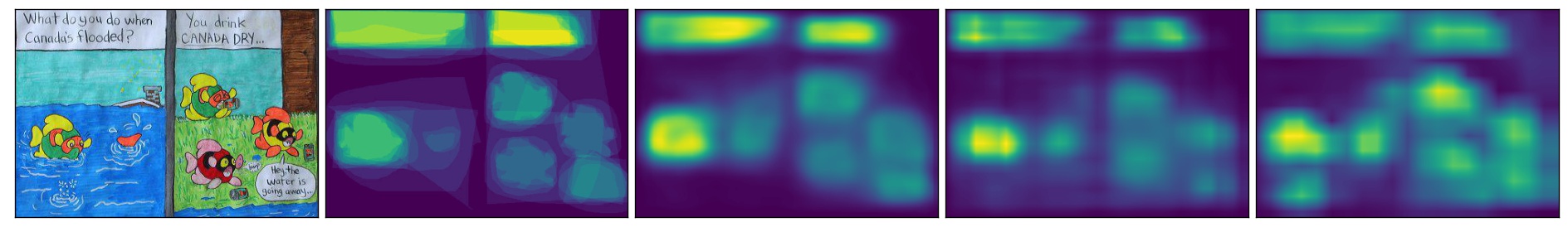}
  \caption{Example images and model predictions from the Imp1k dataset. (a) In this advertisement, the brand is rated as the most important part of the design by human annotators (ground truth). (b) For this webpage and the advertisement above, the natural image saliency model (SAM) over-focuses on the faces. (c) A movie poster, where the prior state-of-the-art importance model (B.~et al.~\protect\cite{bylinskii2017learning}) over-predicts all the text elements as having similar importance. (d) A mobile UI, composed of text and images in equal parts, where SAM focuses on the images and B.~et al. distributes importance evenly. In all the above examples, UMSI correctly detects the most important design elements, regardless of their spatial location on the image. (e) A challenging infographic for all the models, where correctly predicting importance requires an understanding of the data presented. (f) A drawing that does not belong to any other design class, but nevertheless generates correct predictions of importance by the first two models.
  }~\label{fig:fullresults}
\end{figure*}

Example predictions can be found in Fig.~\ref{fig:fullresults}. Training on both natural images and graphic designs allows our model to correctly distribute importance in images that contain different amounts of text and visuals. In these cases, a saliency model might over-predict visual regions, whereas an importance model trained only on posters and ads (such as~\cite{bylinskii2017learning}) struggles to generalize to more complex designs. 

\subsection{Saliency prediction}
We also evaluated our model on the SALICON test set~\cite{jiang2015salicon}. On saliency, our model performs comparably to state-of-the-art. UMSI obtains a CC of 0.782 and KL of 0.341, 
compared to a CC of 0.811 and KL of 0.324 
for SAM \cite{cornia2018predicting}. Qualitatively, we observe very similar patterns as SAM on the output heatmaps: our model correctly detects people and faces, and identifies elements of high contrast (see Figs.~\ref{sam_vs_byl_vs_UMSI} and \ref{fig:flow_of_design}a).

The ability of our model to generalize to natural images and graphic designs alike make it usable within a typical design workflow.  As a proof-of-concept, we generated some designs with a SALICON image as background, and collected importance annotations on designs corresponding to different stages of the design process. On natural images, our model predicts accurate saliency maps. As the designs become more complex, our model switches to distributing importance across the image and design elements (Fig.~\ref{fig:flow_of_design}). We also observe a diminished text bias, as our model accurately recognizes the importance of secondary design elements over text. We note that excessive text bias was one of the limitations of \cite{bylinskii2017learning}.

\subsection{Classification}

Our classification module achieves an average accuracy of 95\% when determining which of 5 classes the designs from Imp1k belong to (per-class accuracy scores are reported in Fig.~\ref{fig:mean_trends}). From Fig.~\ref{fig:mean_trends}, we see that our model has learned to capture general trends of importance that differentiate the 5 design classes. We include some interesting failure cases of our model, when it misclassifies designs, in Fig.~\ref{fig:missclass}.  

We evaluated the contribution of our classification module to importance prediction (Table \ref{tab:Imp1k}). To do so, we trained an additional two versions of UMSI, one without the classification module at all (UMSI-nc), and one where the classification module does not feed directly back into the importance prediction (as in Fig.~\ref{model_arch}), but branches off after the ASPP module, and still affects the features learned by both the encoder and ASPP module (UMSI-2stream). Both alternatives to our architecture performed worse, where not having a classification stream at all (UMSI-nc) affected the scores most.

\section{Applications}

In this section we present two new applications of our importance prediction model. These are proofs-of-concept intended to demonstrate how such a model can help support iterative design and automate some common design workflows. 

\subsection{Model-assisted interactive design}
Previous work used importance prediction within a design tool as passive feedback in the form of real-time importance visualization \cite{bylinskii2017learning}.  
Here we consider a use case where the user can more actively engage with the importance model during the design process to receive design suggestions. 
We developed a bare-bones prototype with minimal design support to evaluate the ability of our model, when coupled with an optimization procedure, to actively adjust a design according to user-imposed constraints.

The workflow allows a user to manipulate design elements on a canvas, and to receive immediate feedback about the predicted importance of each element, similar to the visualizations in \cite{bylinskii2017learning}. However, different from~\cite{bylinskii2017learning} is a new ability to directly interact with the importance scores of each design element, allowing the user to specify constraints to increase or decrease the importance of design elements.

 The application then depends on an optimization procedure to generate new design variants, which are scored by the importance model. For this demo, we chose a genetic algorithm that makes adjustments to the visual arrangement (scale and location) of the design elements and selects adjustments that reduce the gap between the current predicted values and the target values specified by the user. As the optimization algorithm itself is not a contribution of this paper, we leave its implementation details to the Supplemental Material.

\begin{figure}
\centering
  \includegraphics[width=1\linewidth]{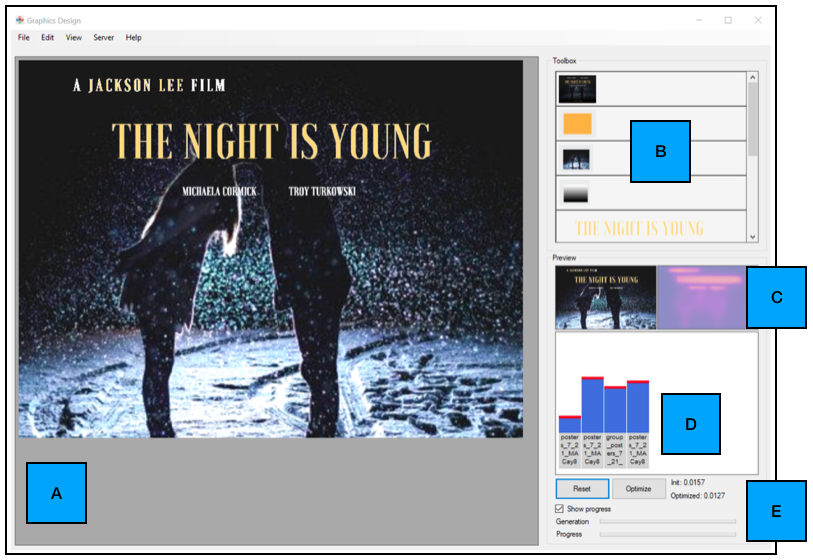}
  \includegraphics[width=1\linewidth]{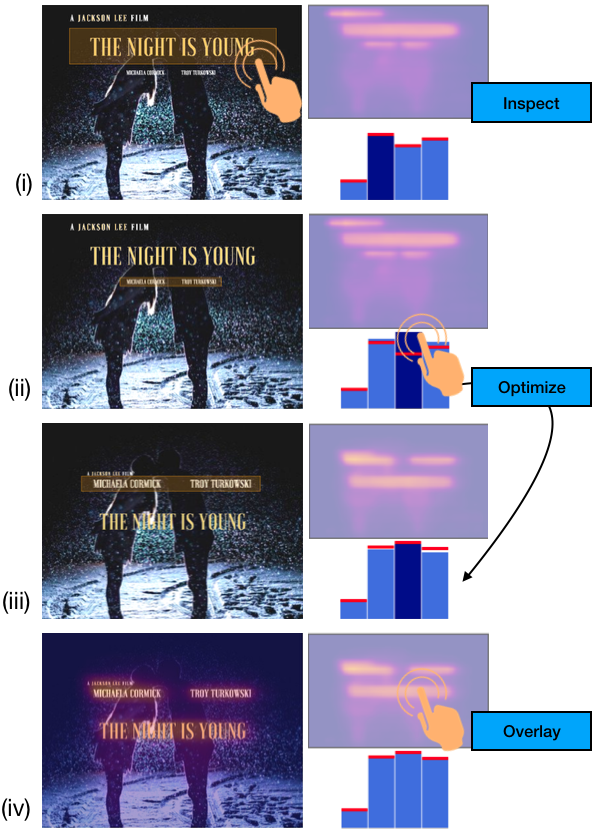}
  \caption{At the top we demonstrate our model-assisted interactive design UI: A) Canvas onto which a user can drag a vector graphics file. B) A set of layers that can be used to compose design elements together. C) A preview of the interactively-computed importance map for the current design. D) An interactive bar plot listing all the design elements and their predicted importance scores. E) A progress bar for the optimization procedure launched when users adjust the interactive bar plot values. Below the UI is a sample interaction sequence: (i) Clicking on a design element highlights the element's predicted importance score in the interactive bar graph. (ii) Making adjustments to the bar graph launches an optimization procedure that automatically adjusts the design (iii) until the target importance values are reached for all the elements. (iv) Selecting the importance heatmap overlays it on the final design.}~\label{fig:ui_interface_seq}
\end{figure}

\textbf{UI design.}
We developed a simple UI for this task which allows drag-and-drop placement of graphical elements (Fig.~\ref{fig:ui_interface_seq}A-B). 
A real-time (TCP network-based) interface between our front-end UI and our back-end importance model allows for remote computation on GPU-enabled computing infrastructure. As users manipulate the design, continuous execution of the importance model in the back-end allows real-time updates of the importance predictions, visualized as a heatmap (Fig.~\ref{fig:ui_interface_seq}C). 
We also compute the individual importance of all the design elements by averaging the importance values inside the mask of each design element. 
We plot these element-wise importance values as an interactive bar plot (Fig.~\ref{fig:ui_interface_seq}D). 
The user can then set individual target importance values for each design element, by adjusting the level of the respective bar in the plot.  
This interaction triggers an optimization procedure that tries to reduce the discrepancy/gap between the current predicted and target importance values (Fig.~\ref{fig:ui_interface_seq}E). 

\textbf{User studies.} We evaluated the ability of our importance model to accurately guide the optimization procedure towards design variants matching user-specified constraints. Using 6 initial designs, we selected 3 elements per design that we wanted to separately increase the importance of. Specifically, we used the interactive bar plot within our UI to constrain the importance of the chosen design element to have maximal importance in the final design. We then launched the optimization procedure for 5 consecutive runs, to produce 5 possible design variants that meet the specified constraints. We produced a total of 90 design variants = 6 (initial designs) x 3 (separate constraints) x 5 (runs). We then launched the ImportAnnots UI to collect importance maps (25 participants worth of annotations per design) for all 90 design variants, as well as for the 6 initial designs. We compared these ground truth importance maps to the ones predicted by our model and used by the optimization procedure to produce each design variant. The average CC score across all 90 variants was 0.928 (where the upper bound of CC is 1.0), indicating a high correspondence between the model and ground truth. On average, across the 6 designs x 3 constraints per design, the optimization algorithm succeeded in increasing the importance of the target element in 87\% of the runs. 

\begin{figure}
\centering
  \includegraphics[width=0.4\columnwidth]{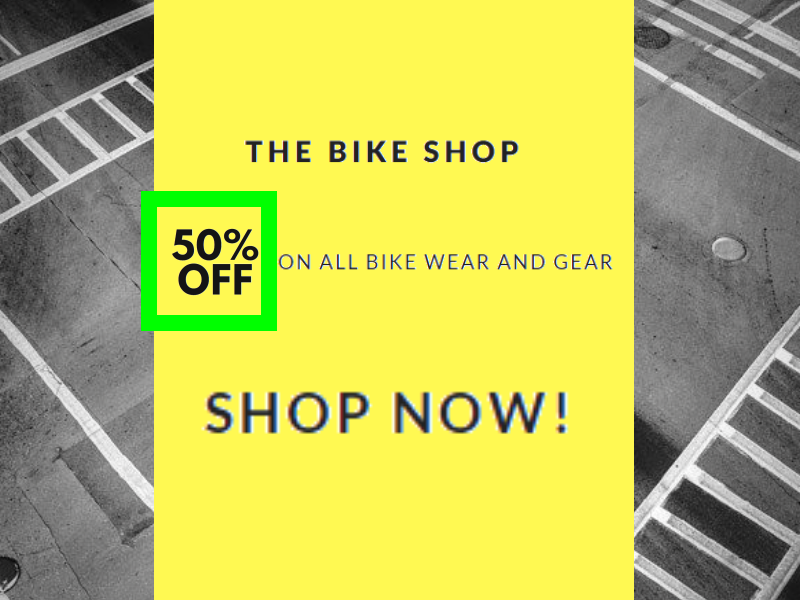}
  \includegraphics[width=0.4\columnwidth]{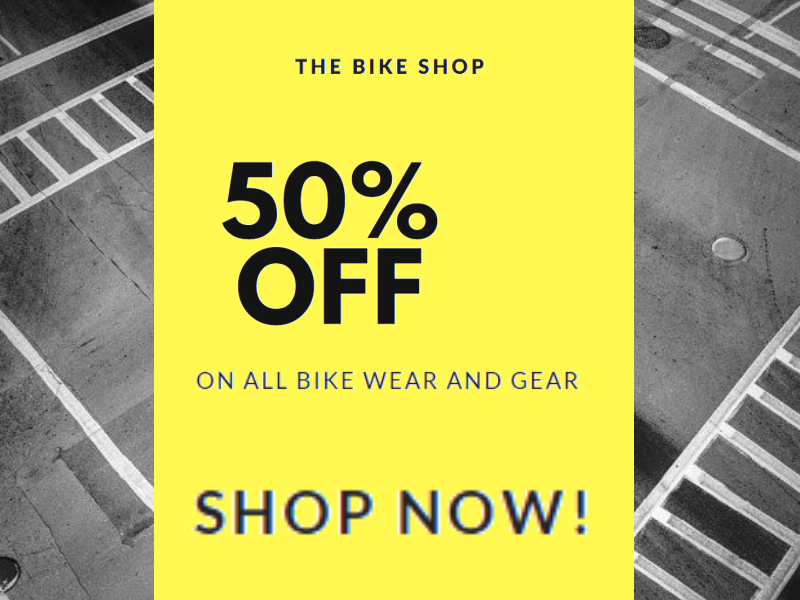}
   \includegraphics[width=0.4\columnwidth]{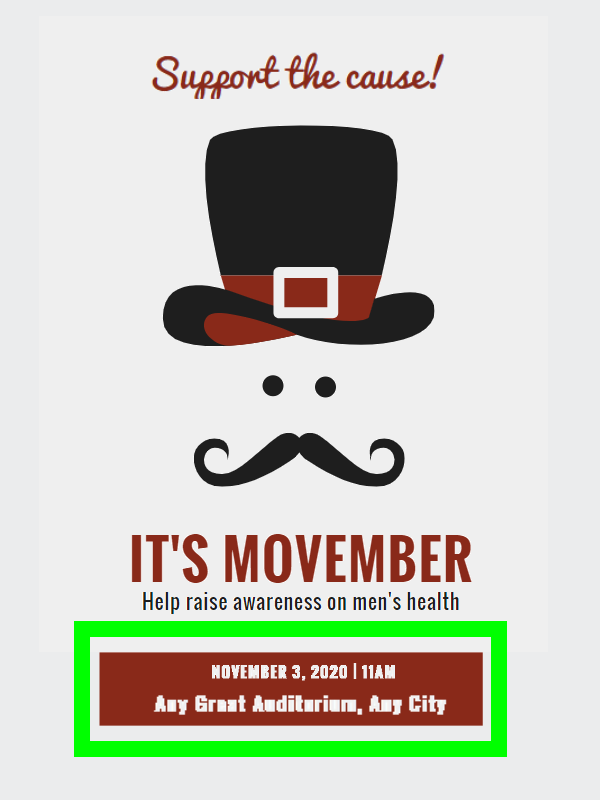}
 \includegraphics[width=0.4\columnwidth]{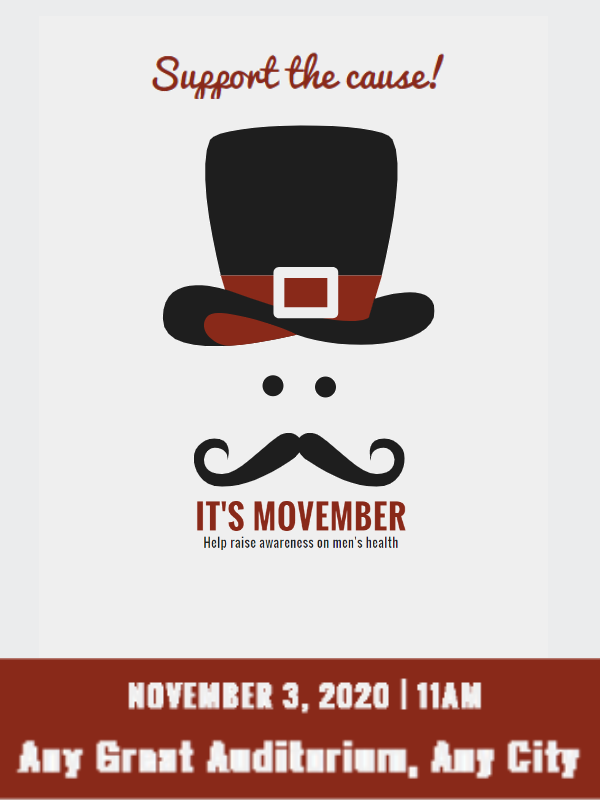}
  \caption{Model-assisted design examples. Left: initial designs. Right: designs after optimization. We maximize the importance of elements marked in green. The algorithm accomplishes this through spatial rearrangement (top) and/or changing proportions (bottom). Changing target importance of one element commonly also affects neighboring elements to preserve their importance levels as closely as possible while giving room for the other element to gain importance.}~\label{fig:ex_ui}
\end{figure}

\textbf{Discussion.} Some example results are provided in Fig~\ref{fig:ex_ui}. The top example contains a simple advertisement where a user wants to emphasize the discount. Our application automatically rearranges the elements such that the design element selected is enlarged and takes a more central position.
 In the bottom example, a user wants to put more emphasis on the location of the advertised event, while also decreasing the prominence of the event title. In the automatic redesign, text about the event time and date is separated from the location, creating more room to enlarge the relevant information. 
 
 In this section we provided a means by which our importance model can be coupled with a separate optimization procedure to offer design suggestions in an interactive design tool. The design tool itself is bare-bones and does not optimize for balance, symmetry, or other aesthetic properties, and so the returned results are not guaranteed to be good quality designs. Despite this, we were able to guarantee, with 87\% success rate, that the importance objectives specified by the user were met, in accurately increasing the importance of specified design elements. These initial investigations show promise for the utility of an importance model in future design tools. \zoya{How best to incorporate AI and other forms of automatic assistance and collaboration in interfaces is a substantial open problem for the field.} 

\subsection{Model-assisted design reflow}

\begin{figure*}
\centering
  \includegraphics[width=1\textwidth]{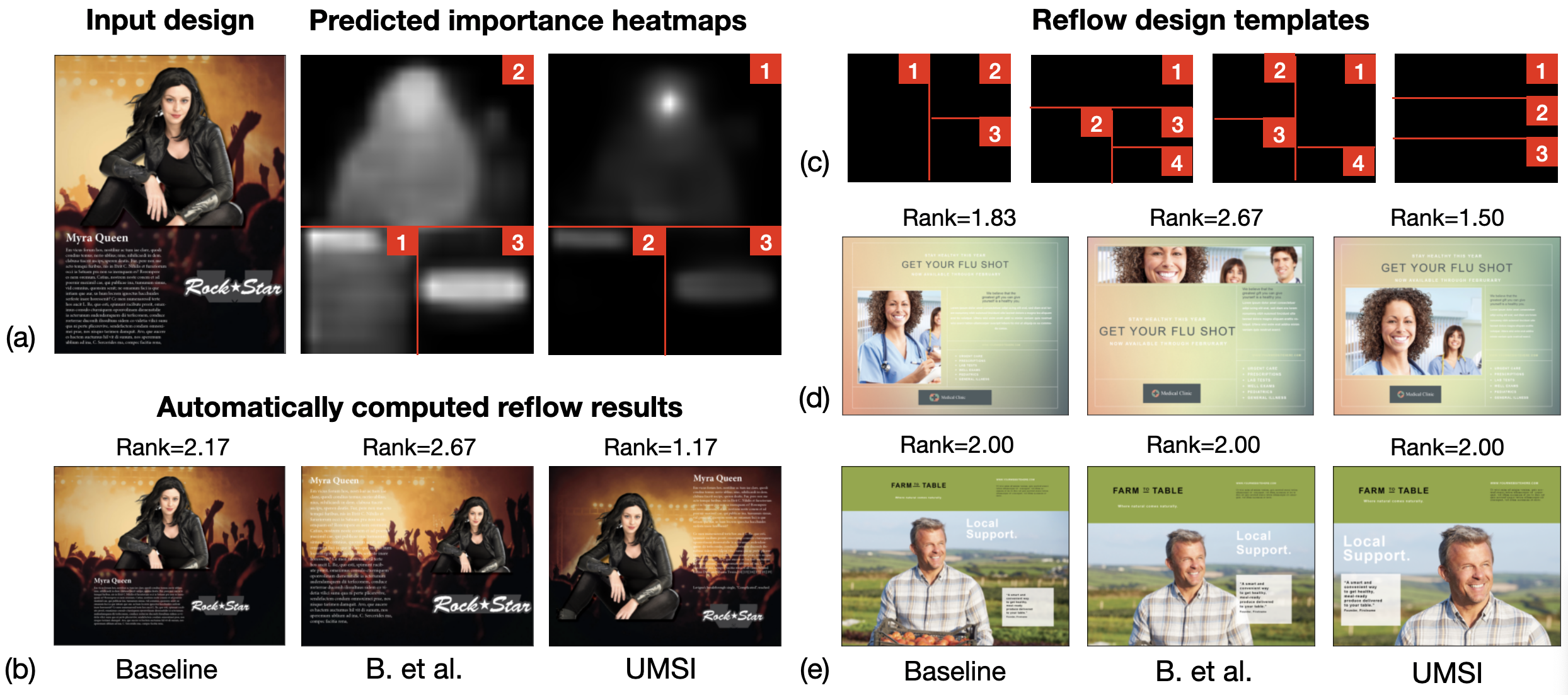}
  \caption{Model-assisted design reflow. (a) An input design and importance heatmaps predicted by \zoya{B. et al.}~\protect\cite{bylinskii2017learning} and our UMSI model, respectively. These heatmaps are used to rank the individual design elements by importance (numbered insets). (b) During reflow, design elements are then moved and rescaled by being mapped to placeholders in retrieved templates. (c) Template examples. The first template was used for the results in (b), while the second template was used in (d) and (e). (d,e) More examples of reflow results, comparing a baseline \zoya{(described in text)}, to importance-aided reflow with the model \zoya{B. et al.}~\protect\cite{bylinskii2017learning} and with UMSI. \zoya{Crowdsourced ranking scores are included above each variant (the lower, the better).}}~\label{fig:reflow}
\end{figure*}

We also investigated the benefits of using importance prediction 
as a back-end mechanism for creating different sized variations of an input design. This is required for adapting designs to alternative form factors and devices.
We propose using the importance scores of design elements computed from the importance heatmaps to guide repositioning and rescaling of the elements when reflowing a graphic design (Fig.~\ref{fig:reflow}). 

We worked with the developers of a commercially-available layout application to integrate our importance models into design reflow.
The application lets users create or upload their existing designs and perform basic operations, like placing, scaling, and grouping graphical elements. Once the user is content with the initial design, the design is sent as an image to our back-end importance model. The predicted importance heatmap is then used to rank the graphic elements by importance, by computing the average importance value over each element's extent. Our proposition is that by preserving the relative importance values of all the design elements in the reflow result, we maintain the designer's intention in allocating attention across design elements.

\textbf{Reflow algorithm.} Given an input design and a new target design size (Fig.~\ref{fig:reflow}a), the reflow algorithm selects a new position and scale for each design element (Fig.~\ref{fig:reflow}b). 
For designs of a few different sizes and numbers of elements, we manually composed templates indicating the importance rank of each element in that design (examples in Fig.~\ref{fig:reflow}c). We composed these templates based on professional designs of similar sizes and number of elements. The templates are stored in the reflow application. At test-time, we retrieve the template matching the input design in number of design elements, and with similar aspect ratio to the target design size. 
Next, we map each element from the initial design to a placeholder in the template with a matching importance rank. This step preserves the importance ranks of the elements from the original design.

\begin{figure}[h!]
  \centering
  \includegraphics[width=1.0\columnwidth]{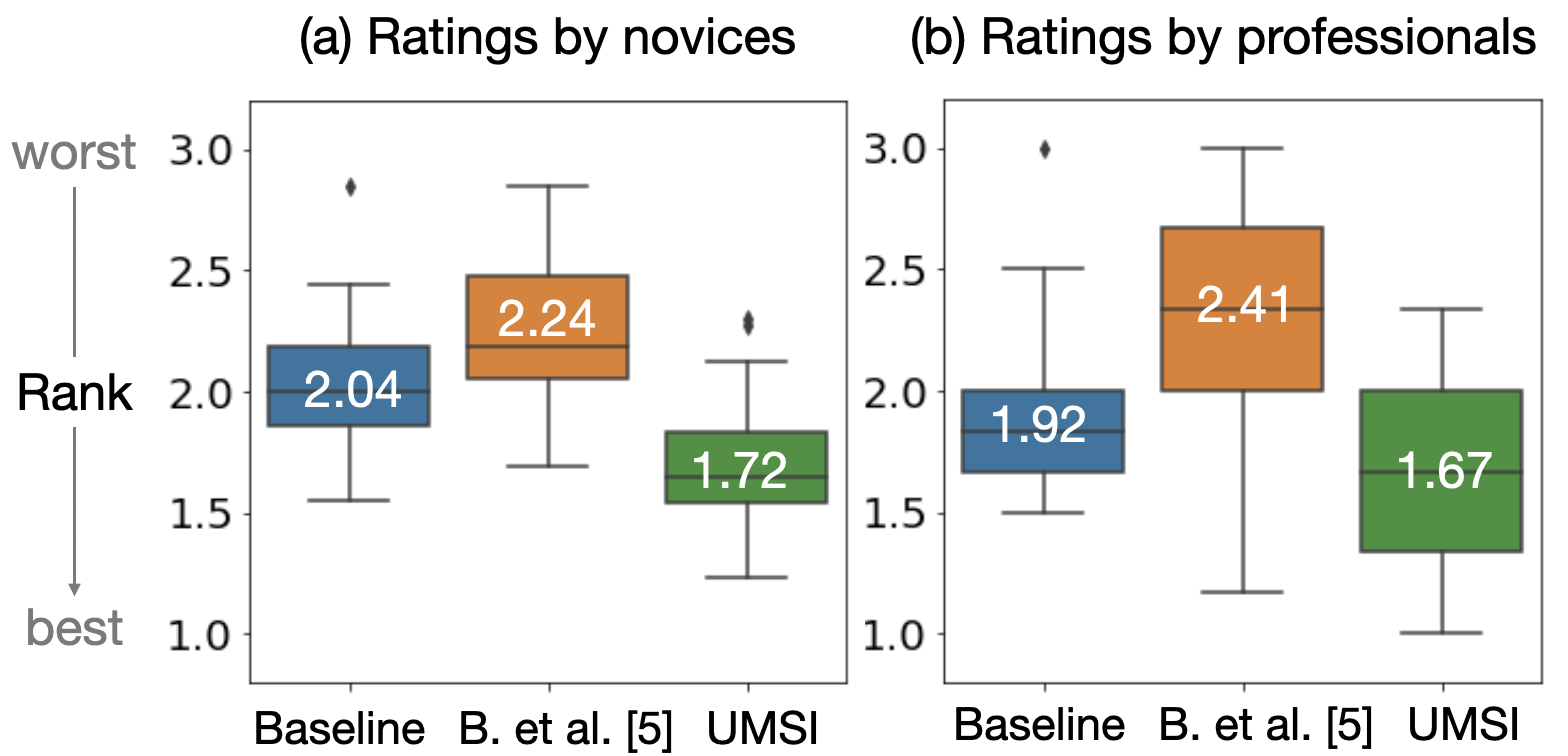}
  \caption{User study results for model-assisted design reflow. We compare a baseline (no importance model used) to reflow aided by a previous importance model~\protect\cite{bylinskii2017learning} and UMSI, respectively. All 3 reflow variants were ranked from 1 (best) to 3 (worst) for 17 different designs by an average of 23 novice crowdworkers (a) and 6 professional designers (b). UMSI-aided design reflow results were rated significantly better than the alternatives by both participant pools.} 
  \label{reflow-user-study}
\end{figure}

\textbf{User studies.} To evaluate whether using our importance model at the back-end of a layout application indeed improves the final reflow results, we ran two user studies, with crowdworkers and with professional designers, respectively. For 17 different graphic designs, we provided study participants with three automatically-computed design variants in an alternative aspect ratio. Participants were asked to rank the variants from 1 (best) to 3 (worst), based on personal judgement without further guidance. These variants corresponded to: (i) a baseline implementation of reflow without using importance\footnote{An existing constraint-based engine that was the default in the commercially-available layout application we used.}, and results from the importance-aided reflow algorithm described above, using: (ii) the previous state-of-the-art importance model~\cite{bylinskii2017learning} and (iii) our UMSI model. The ordering of the variants was randomized across designs and participants. We ran the study on Amazon's MTurk, and recruited 100 participants, each presented with 9 randomly-sampled designs and 3 repeat designs, and compensated \$1.50. To ensure high quality data, we only kept the participants that consistently ranked the variants on 2/3 of the repeat designs. The data of 43 participants (29 male, most in their 20s and 30s) were used in the resulting analyses. Based on an average of 23 participant rankings for each of the 17 designs, we found that the reflow results using the UMSI model performed significantly better ($p<0.05$, Bonferonni-corrected) than the other two reflow variants (Fig.~\ref{reflow-user-study}a). The difference between importance-based reflow using~\cite{bylinskii2017learning} and the baseline was not significant, however. We repeated the task with 6 professional designers recruited from personal networks (3 male, between 25-45 years old, with a variance of 3-20 years of design experience). Each professional was asked to rank the three design variants for all 17 designs, and was compensated \$5.00. Based on their average rankings of the design variants, we confirm that the reflow results using the UMSI model performed better than the other two reflow variants (Fig.~\ref{reflow-user-study}b). Examples of reflow results from the three methods compared can be found in Fig.~\ref{fig:reflow}b,d,e, along with their average ranks from the MTurk study. 

\textbf{Discussion.} While we evaluated the benefit of guiding reflow using an importance model within a particular design tool, we believe that the observed gains in the quality of reflow results are encouraging. Future reflow applications could incorporate importance as an additional constraint (i.e., by preserving the importance ranks of design elements in the original and redesign versions), when combined with balance, symmetry, and other common aesthetic metrics. Indeed, one of our professional designer participants, who was naive to the purpose of the study and the sources of the retargeting results, described that what makes up a good reflow result is \emph{``the hierarchy of elements, making sure we read the most important thing first."} \peter{While our approach only modifies element size and position, future work could consider colors, fonts, and other features to affect visual importance during reflow.}

Notably, as seen from our user study, we achieved improved reflow results with our proposed UMSI importance model, but not with the prior state-of-the-art model~\cite{bylinskii2017learning}. The UMSI model generalizes better to more diverse and complex designs and its predictions are less biased towards text (Fig.~\ref{fig:reflow}b,d) - properties that make our importance model more amenable to the reflow application. In Fig.~\ref{fig:reflow}e, we see an example where the reflow results of all methods were scored comparably, with the key difference being that each successive model zoomed in more on the man's face. UMSI was trained on natural images in addition to graphic designs, and as a result has learned that faces are very salient to human observers, and should be prioritized. The baseline model produces a result that is most like a crop of the input design. The original designs and more reflow results can be found in the Supplemental Material.

\section{Conclusion}     

In this paper, we presented the first Unified Model of Saliency and Importance, capable of approximating a notion of human attention on natural scenes and graphic designs alike. We showed that our model is competitive with state-of-the-art models that have been specialized for different tasks: predicting natural scene saliency, and graphic design importance, respectively. Not only can our model generalize to these broadly different input modalities, it can also provide accurate fine-grained predictions on different design classes, which makes two new interactive design applications possible. We presented a model-assisted design application that automatically adjusts the elements in a vector design to meet user-specified importance constraints; and a graphic design reflow application that automatically adjusts the locations and sizes of design elements to fit to new aspect ratios. 
The model architecture presented in this paper is not limited to the design classes in Imp1k. The training procedure can be adapted, given a new training set, to any number of additional design classes. This opens up the possibility of continually improving the current UMSI model to make it a one-stop shop for predicting attention on any images, natural and designed.

\balance{}

\bibliographystyle{SIGCHI-Reference-Format}
\bibliography{sample}

\end{document}


\title{\plaintitle}

\numberofauthors{1}
\author{%
 \alignauthor{
    Camilo Fosco\textsuperscript{1},
    Vincent Casser\textsuperscript{1},
    Amish Kumar Bedi\textsuperscript{2},\\
    Peter O'Donovan\textsuperscript{2},
    Aaron Hertzmann\textsuperscript{3},
    Zoya Bylinskii\textsuperscript{3}, \\
   \affaddr{\textsuperscript{1}MIT}
   \affaddr{\textsuperscript{2}Adobe Inc.}
   \affaddr{\textsuperscript{3}Adobe Research}\\
   \email{\{camilolu, vcasser\}@mit.edu}
   \email{\{ambedi, podonova, hertzman, bylinski\}@adobe.com}
   }
 }

\maketitle


\section{Model-assisted interactive design}

Once a user triggers an interaction with the bar graph of importance scores, we automatically generate $N=100$ new design variants (in the background). In a given variant, every graphical element is adjusted with a probability of $p=0.5$ via positional offsets and scaling factors. We evaluate each design variant by running our importance prediction model and measuring the mean-squared-error (MSE) between the predicted and target importance scores. We add penalty terms for overlapping elements. We keep the top 25 of designs with the lowest MSE. We generate a new set of 75 designs by ``genetic crossover": combining positional offsets and scaling factors for individual design elements across pairs of designs from the top 25, to produce a new ``population" of 100 designs. The idea of crossover here is to allow for the effective combination of positive aspects from different variants. We repeat this genetic design breeding process for 20 epochs. After every epoch, we update the canvas with the best design so far. 

\section{Model-assisted design reflow}

All 17 designs and their corresponding reflow variants from our user studies are shown in Figures 1-5.

\begin{figure*}
\centering
  \includegraphics[width=1\linewidth]{figures/reflow/1_.png}
  \includegraphics[width=1\linewidth]{figures/reflow/2_.png}
  \includegraphics[width=1\linewidth]{figures/reflow/3_.png}
  \includegraphics[width=1\linewidth]{figures/reflow/4_.png}
  \caption{}~\label{fig:reflow}
\end{figure*}

\begin{figure*}
\centering
  \includegraphics[width=1\linewidth]{figures/reflow/5_.png}
  \includegraphics[width=1\linewidth]{figures/reflow/6_.png}
  \includegraphics[width=1\linewidth]{figures/reflow/7_.png}
  \includegraphics[width=1\linewidth]{figures/reflow/8_.png}
  \caption{}~\label{fig:reflow}
\end{figure*}

\begin{figure*}
\centering
  \includegraphics[width=1\linewidth]{figures/reflow/9_.png}
  \includegraphics[width=1\linewidth]{figures/reflow/10_.png}
  \includegraphics[width=1\linewidth]{figures/reflow/11_.png}
  \includegraphics[width=1\linewidth]{figures/reflow/12_.png}
  \caption{}~\label{fig:reflow}
\end{figure*}

\begin{figure*}
\centering
  \includegraphics[width=1\linewidth]{figures/reflow/13_.png}
  \includegraphics[width=1\linewidth]{figures/reflow/14_.png}
  \includegraphics[width=1\linewidth]{figures/reflow/15_.png}
  \includegraphics[width=1\linewidth]{figures/reflow/16_.png}
  \caption{}~\label{fig:reflow}
\end{figure*}

\begin{figure*}
\centering
  \includegraphics[width=1\linewidth]{figures/reflow/17_.png}
  \caption{}~\label{fig:reflow}
\end{figure*}

\section{Additional Training Details}
We train with KL and CC losses, with coefficients of 10 and -3, A binary cross-entropy loss with a weight of 5 is used for the classification submodule. We obtained these coefficients through grid search and 3-split cross validation, testing values of 1, 3, 5, 10, 15 for each loss (negative values for CC). The learning rate and dropout rate was similarly defined, testing 5 values with 3-split cross validation. The definition of the architectural modules follows insights from previous literature, including image classification \cite{chollet2017xception, szegedy2016rethinking} and image segmentation work \cite{chen2018deeplab, chen2017rethinking}.

\bibliographystyle{SIGCHI-Reference-Format}
\bibliography{sample}